# Toward a Thinking Microscope: Deep Learning in Optical Microscopy and Image Reconstruction

Yair Rivenson and Aydogan Ozcan

Electrical & Computer Engineering Department, University of California, Los Angeles, CA, 90095, USA

Bioengineering Department, University of California, Los Angeles, CA, 90095, USA

California NanoSystems Institute, University of California, Los Angeles, CA, 90095, USA

http://innovate.ee.ucla.edu/welcome.html

http://org.ee.ucla.edu/

**Abstract**: We discuss recently emerging applications of the state-of-art deep learning methods on optical microscopy and microscopic image reconstruction, which enable new transformations among different modes and modalities of microscopic imaging, driven entirely by image data. We believe that deep learning will fundamentally change both the hardware and image reconstruction methods used in optical microscopy in a holistic manner.

Recent results in applications of deep learning [1] have proven to be transformative for various fields, redefining the state of the art results achieved by earlier machine learning techniques. As an example, one of the fields that has significantly benefited from the ongoing deep learning revolution is machine vision, with landmark results that enable new capabilities in autonomous cars, fault analysis, security applications, as well as entertainment. Some other fields which have recently achieved technological breakthroughs by using deep learning include voice recognition and speech translation, which are transforming the ways that we communicate with each other and devices. This ongoing deep learning revolution is the result of a perfect storm that brings together multiple key factors, i.e., access to enormous amounts of data that have been generated by digital sensors and consumer devices, and the availability of powerful and yet cost-effective computational tools, both at the hardware and software level, making large-scale statistical learning a tractable challenge. These learning tasks can be crudely divided into supervised (where known data labels are being used as feedback to train a statistical model) and unsupervised learning (where the learning is dictated by a policy of rewards, without label-based feedback).

Here in this article, we focus on supervised learning and the applications of the state-of-art deep learning approaches on optical microscopy and microscopic image reconstruction and image transformation in general. Supervised deep learning approaches, where the data labels are known or acquired in advance, have been extensively used in e.g., medical imaging applications, to diagnose diseases (e.g., skin cancer), determine tumor margins, or in general classify and segment different regions of the sample field-of-view into various pre-determined classes, among others. Unlike this existing plethora of main stream applications of deep learning, in this article we focus on another very exciting emerging opportunity: to reconstruct microscopic images and transform optical microscopy images acquired with a certain imaging system into deep learning-enhanced new images that exhibit e.g., improved resolution, field-of-view, depth-of-field and/or aberration correction, statistically matching the images that would be expected from a higher-end imaging system. This specific direction of deep learning-enabled image reconstruction or image transformation is especially suitable for optical microscopy since it provides perfect control of all the imaging parameters needed for generating high-quality training image data, with nanoscopically aligned image pairs, between e.g., a low resolution or aberrated microscope and a higher resolution diffraction-limited one. Compared to photography or macro-scale imaging in general, optical microscopy provides a highly-controlled and repeatable platform, where deep learning can show its ultimate performance under perfectly aligned and labeled training image pairs, helping us focus solely on the image reconstruction or transformation that the neural network is being trained for.

Following the training image data collection, a model that learns the statistical image transformation from an input distribution *X* to the desired (i.e., enhanced) distribution *Y* is trained with the matched pairs of training data, containing typically thousands of image patches. Before such a model can be trained, its architecture needs to be determined and optimized. For deep learning, the architecture is typically composed of cascaded convolutional layers, where each convolutional layer contains multiple convolution kernels and bias terms (trainable constants) that map the input of that layer into feature maps, or channels. Generally, an extended number of convolutional filters should be able to generate more complex network models, and in theory, can learn more complicated tasks with larger datasets. However, the price to pay for this, in general, is an increase in the training and inference times as well as a potential caveat for overfitting to training data. In a deep network's hidden layers (i.e., the layers between the input and the output layers), the convolutional layers are followed by activation functions (i.e., neurons) that introduce non-linearity to the deep network, which helps the network to generalize its inference to a broader class of functions/tasks. A popular selection of the neuron function is the rectified linear unit (ReLU), which outputs *w* = *max*(0,*z*), where *z* and *w* are the input and the output, respectively, and *max* refers to the maximum operation. Other activation functions are also being frequently used in different architectures and they are chosen according to the task at hand.

The training of the network based on labeled data can be considered as an optimization problem, where the network attempts to optimize its output with respect to the gold standard target labels, given a user predefined cost (or loss) function. As an example of a cost function, one can choose to minimize the energy difference (e.g., the mean squared error) between the network's output images, $\hat{Y}$, and their corresponding gold standard labels, acquired by e.g., a higher-end microscope that we train the neural network to mimic. A typical network training procedure is outlined in Fig. 1. For each pair of training images, the amount of error is calculated through the predefined cost function, and this error is propagated back within the network to update all the weights and biases in the direction that minimizes the error or the cost. This process is also known as the error back-propagation in deep learning, and through this iterative algorithm, the model adjusts its coefficients to satisfy the given target labels and minimize cost function constraints. The available training image data are being used multiple times in this iterative learning process to fine tune the model's inference success, and once the entire training dataset is used during these iterations, we complete an epoch; typically tens to thousands of epochs are used to complete the training phase, depending on the complexity of the network and the task that it is charged with. After this phase, the network is ready to be blindly tested with new data sets that were not part of the training or validation sets.

One of the most exciting aspects of deep learning-enhanced optical microscopy and image transformations is that, once the network is trained, the inference is non-iterative and very fast to compute without the need for any parameter search or optimization. In this sense, compared to other solutions to inverse problems in optics, using e.g., deconvolution, convex optimization or compressive sampling/sensing techniques, a trained deep neural network, once optimized, presents significant advantages in terms of computation speed and inference time, even using modest computers and processors. In fact, with the emergence of new processor architectures that are specifically optimized for deep learning, this important advantage will take another level of significance, helping us perform the inference tasks of neural networks in real time, even with mobile phones and other low-end consumer devices.

It is indeed true that the network training process can be quite lengthy (from a few hours to more than a day), depending on e.g., the size of the training data, available hardware and model complexity; however, once the model is trained it remains fixed and can also be used to warm start new models, when new data become available or new tasks are required. This process of applying one trained neural network onto a new task with new data is known as transfer learning, which significantly accelerates a new network's training time.

One of the first applications of deep learning to enhance optical microscopy images was demonstrated using bright-field microscopy [2]. In this work, histochemically stained lung tissue sections were imaged using a bright-field microscope with a 40×/0.95NA objective lens to obtain lower resolution (LR) images of specimen, and a 100×/1.4NA oil-immersion objective lens was used to obtain the corresponding high-resolution (HR) labels or gold standard images, to be used for training a convolutional neural network. A deep neural network architecture was then designed to transform LR images (used as input) into enhanced images that match the HR labels. The task that the network was aiming to learn in this case can be thought as predicting the pixel values of the HR image, given the LR input. Therefore, an important step prior to training was to correctly align, or register, the LR and HR training images with respect to each other; this enforces the deep network to solely learn the LR-to-HR transformation, rather than some arbitrary affine transformation between the input and output images. Following the accurate alignment of the images, the network model can be trained with the matched LR and HR image pairs. One key advantage of this deep learning-based image transformation is that, unlike other image enhancement or deconvolution methods, there is no need to have *a priori* information about the object or the image formation process; stated differently modelling of the point spread function, spatial and spectral aberrations, illumination properties or other physical parameters of the imaging system or the object, and their impact on the acquired image do not need to be known or estimated since the neural network uses training image data to inherently learn these details in its multi-dimensional solution space.

Following its training, this bright-field microscopic image enhancement network was blindly tested on Masson's Trichrome stained lung tissue sections, which were taken from a different patient. The network output, in response to LR input images, super-resolved the blurry and distorted features in the input images, providing similar images to the ones that were acquired with a 100×/1.4NA objective [2]. This trained network was also robust in imaging new types of samples that were not part of the training data (see Fig. 2). For example, the same model was tested on a different tissue type (kidney), which was also stained using the Masson's Trichrome stain. The inference results showed that the network was indeed able to enhance the resolution of the imaged specimen, demonstrating the generalization of its inference capability. Moreover, the same deep network that was trained with Masson's Trichrome stained lung tissue sections was able to super-resolve different tissue types that used a different stain, e.g., breast tissue sections labeled with H&E (Haemotoxylin and Eosin), see Fig. 2. As a further evidence of its generalization, the same network was blindly tested on a resolution test target (Fig. 2), clearly revealing its enhanced resolution and data-driven frequency extrapolation capability [2].

Another interesting feature of this bright-field microscopy network is that it extends the inferred image's depth-of-field. Since the input images are acquired using a lower NA objective, the network learns to enhance all the spatial features which appear in focus in a low-NA image, resulting in an output image with high resolution and extended depth-of-field [2]. Because the input images have significantly larger field-of-view compared to a higher NA objective lens that the network is trained for, the network output images also exhibit increased field-of-view, further enhancing the imaging throughput of the bright-field microscopy system through deep learning. Quite importantly, even using a laptop equipped with a GPU, the network output image can be calculated in less than a second, [2] without any iterations or parameter tuning, potentially enabling real-time performance using more advanced computational resources and parallel computing.

This microscopic image enhancement and the presented LR-to-HR image transformation are not limited only to high-end microscopy equipment. In fact, it was recently demonstrated that deep learning can be used for significantly improving the imaging performance of mobile microscopes [3]. Creation of cost-effective and portable microscopic imaging devices based on e.g., mobile phones has been significant improved in the last few years, with potential impact on global health and point-of-care diagnostics [4–9]. However, such mobile microscopy devices still have various sources of imperfections and aberrations compared to laboratory-grade microscopes that are used in clinical applications. This gap is partially related to various design constraints imposed by the compactness, cost-effectiveness and extremely-large volume manufacturing of mobile phones. To bridge this performance gap between mobile phone-based microscopes and their benchtop counterparts, a recent work used a deep learning model that, in response to a mobile phone input image, rapidly outputs an image aiming to match the image quality acquired by a high-end benchtop microscope [3]. As in the previous example, to train the network thousands of aligned image patches, acquired using the smartphone and benchtop microscopes, were used. Some examples of the network inference results and their comparison to benchtop microscope images of the same samples are illustrated in Fig. 3. These results, once again, emphasize an important feature of deep learning-based microscopic image enhancement: there is no need for numerical or analytical modelling of the spatially- and spectrally-varying aberrations of the mobile microscope in order to set-up an inverse problem based on a forward model. In fact, such a model is rather complicated to establish for a mobile phone-based microscope as the repeatability of a cost-effective and hand-held platform is not high. Therefore, image data-driven training of a deep neural network provides an elegant solution to statistically learn the transformation between LR and/or aberrated input images and the HR labels without any physical modelling of the image formation process. To better illustrate the robustness of this learned image transformation, highly compressed JPEG images (with a file size that is >20-fold smaller per image), captured by the same smartphone microscope, were used as inputs to the same neural network, with very similar inference results, matching the HR labels acquired using a benchtop bright-field microscope [3]. This ability of the network to be able to work with significantly compressed input images is especially important for resource scarce settings that have limited data transmission bandwidth and storage capacity.

In addition to bright-field microscopy, deep learning has also been applied to improve other optical microscopy modalities, including fluorescence microscopy [10–13] and holographic microscopy [14–16]. Especially, the latter exemplifies a unique opportunity provided by deep learning on microscopic image reconstruction, opening up data-driven alternatives to decades-old analytical or iterative image reconstruction methods that are physics-driven. Coherent holographic microscopy has the advantage of indirectly detecting both the amplitude and phase of an object. While coherent holographic imaging systems offer some unique opportunities for label-free analysis of samples, they in general suffer from the "missing phase" problem. This missing phase information at the detector plane means that the numerically formed object image, unless the phase is recovered, will be plagued by artifacts, such as the self-interference noise and the twin image. The latter is especially stronger for in-line holographic microscopy [17–21], where the object wave and the reference wave co-propagate along the same direction. Various phase recovery and holographic image reconstruction approaches have been developed over the years to solve this missing phase problem. Purely algorithmic approaches require prior information on the object function such as its spatial support or sparsity representation. Other approaches involve hardware modifications of the holographic imaging set-up to facilitate measurement diversity and the use of additional measurements (via e.g., multiple angles of illumination, multiple sample-to-sensor distances, phase-shifting, etc.) as physical constraints for phase recovery and image reconstruction.

As an alternative to these physics-driven phase recovery and holographic image reconstruction approaches, data-driven methods based on deep learning have been recently demonstrated to perform holographic image reconstruction from a single hologram, providing significant savings in both hologram acquisition and reconstruction times [14,15]. One of these deep learning models was trained using numerically back propagated in-line hologram intensities (without the phase information) as the complex-valued inputs to the network, while the target images (i.e., the labels) were reconstructed by an iterative multi-height phase retrieval method that used eight different holographic measurements of the same object, taken at different sample-to-sensor distances, providing physical measurement diversity for accurate phase retrieval [15]. After its training, this phase recovery and holographic image reconstruction network was blindly tested by imaging various samples including breast tissue sections, blood and Papanicolaou (Pap) smears (Fig. 4) using a single hologram measurement in each case, matching the gold standard images that were obtained using eight holograms of the same sample, processed with the iterative multi-height phase retrieval algorithm [15]. This phase recovery network successfully learned to eliminate self-interference and twin image related artifacts that are normally superimposed onto the phase and amplitude channels of the object's image; however, despite all its training data, the network did not learn the physics of wave propagation or hologram formation process as it was not trained for it. In fact, some of the out-of-focus objects (e.g., dust particles [15]) that lied outside of the sample plane were also cleaned/removed by the neural network although they are physical objects that should appear in a physics-driven hologram reconstruction method. This means, instead of providing a solution that is compatible with the wave equation, the network only learned the image transformation ($X \rightarrow Y$) that it was statistically trained for. Another benefit of this deep learning-based holographic image reconstruction approach is its rapid reconstruction time, which was at least 4-fold faster compared to iterative reconstruction approaches [15]. As a further advancement, a recent work has utilized a similar deep learning framework to perform both auto-focusing and phase recovery tasks in a single neural network, which provided significantly larger depth-of-field in holographic imaging while also drastically improving the algorithm time-complexity of holographic image reconstruction in general [16].

In summary, deep learning has been transforming optical microscopy and image reconstruction methods by enabling new transformations among different modes and modalities of microscopic imaging, all driven by image data. We believe that deep learning will become an essential component of modern optical microscopy, fundamentally changing both its hardware and image reconstruction methods in a holistic manner. Especially with the emergence of powerful *unsupervised* learning approaches, which the deep learning field is still at its infancy, the impact of a "***learning and thinking microscope***" will be unprecedented, leading to unique capabilities and enabling applications that are simply not possible with today's optical microscopy technologies.

**Acknowledgements**

The Ozcan Research Group at UCLA acknowledges the support of NSF Engineering Research Center (ERC, PATHS-UP), the Army Research Office (ARO; W911NF-13-1-0419 and W911NF-13-1-0197), the ARO Life Sciences Division, the National Science Foundation (NSF) CBET Division Biophotonics Program, the NSF Emerging Frontiers in Research and Innovation (EFRI) Award, the NSF INSPIRE Award, NSF Partnerships for Innovation: Building Innovation Capacity (PFI:BIC) Program, the National Institutes of Health (NIH, R21EB023115), the Howard Hughes Medical Institute (HHMI), Vodafone Americas Foundation, the Mary Kay Foundation and Steven & Alexandra Cohen Foundation.  Yair Rivenson is partially supported by the European Union's Horizon 2020 research and innovation programme under the Marie Skłodowska-Curie grant agreement No H2020-MSCA-IF-2014-659595 (MCMQCT).

**Figures and Figure Captions**

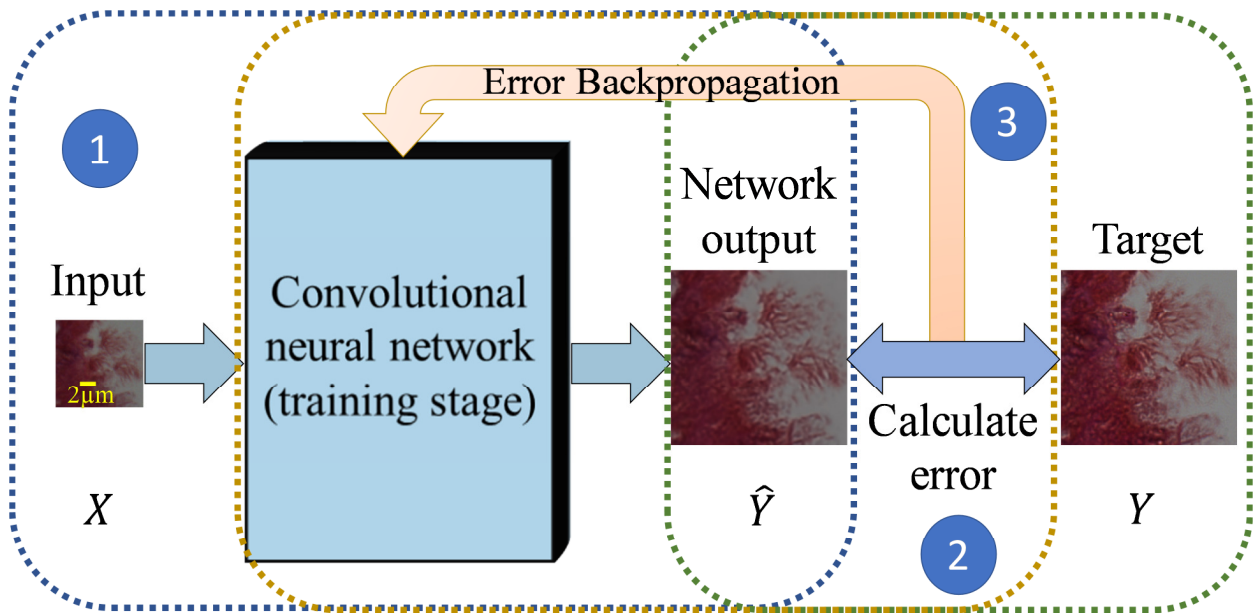

**Fig. 1.** Deep neural network training process. (1) The input is being forward-passed through the network, with the output determined by the current set of weights and biases of the network. (2) The resulting output image is compared to the desired target image (i.e., gold standard), and the error between the two images is calculated, according to a predetermined cost function. (3) Error backpropagation is then used to adjust the weights and biases of the network to minimize the cost function. This process is performed iteratively by using the entire training data in several rounds (i.e., epochs).

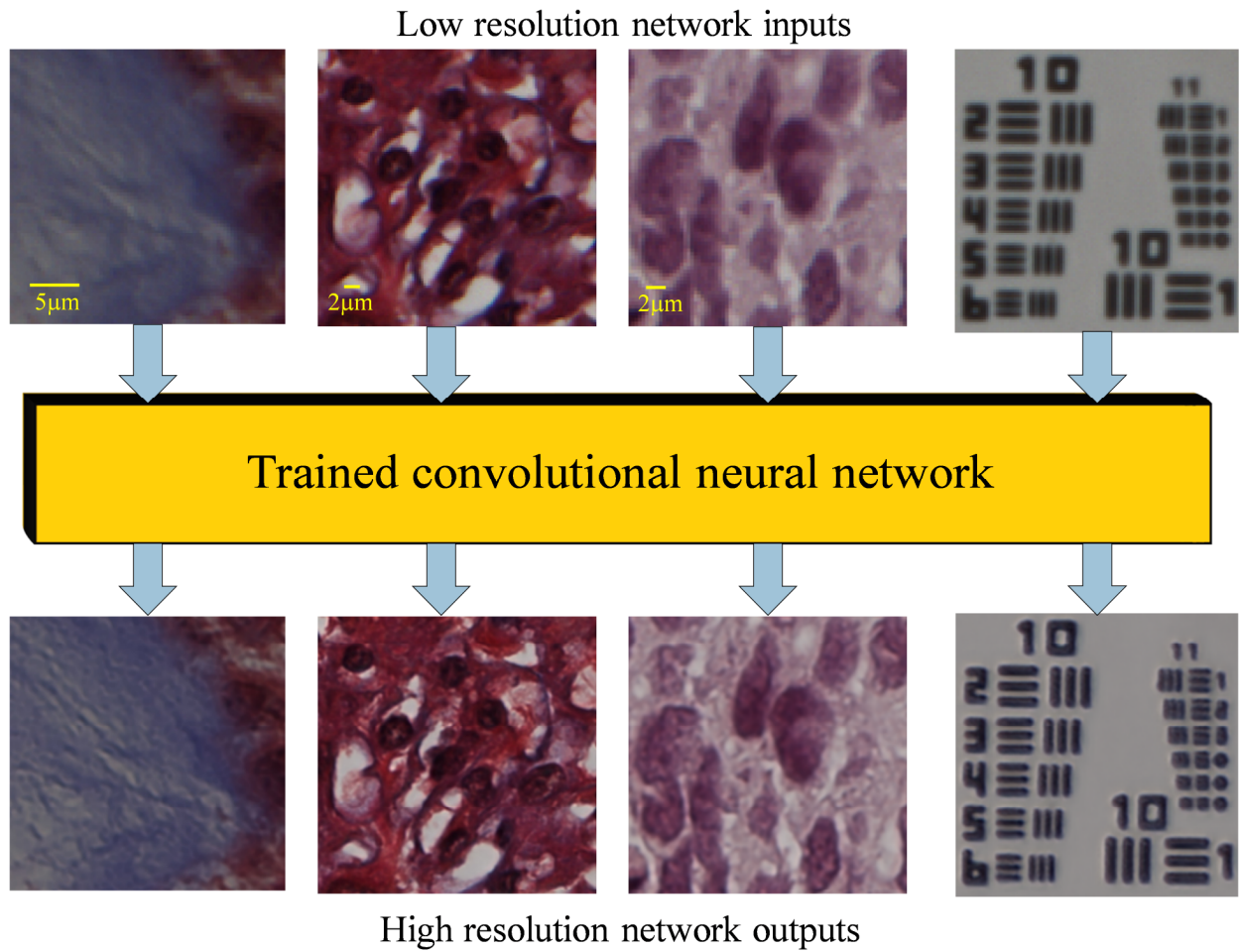

**Fig. 2.** Examples of the network inference using deep learning. Although the neural network was trained with only images of lung tissue samples (left column), its performance proved to be general for other types of samples, including kidney and breast tissue sections and a resolution test target (middle and right columns), which were never seen by the network during its training.

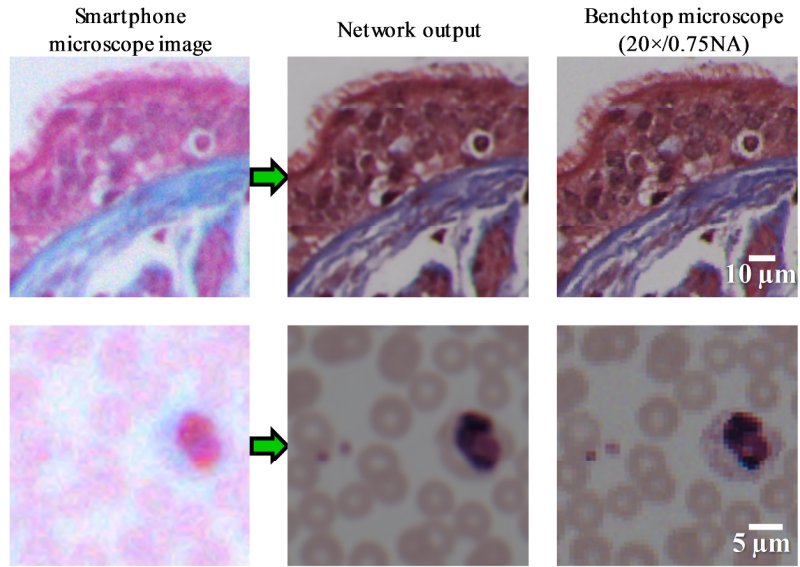

**Fig. 3.** A trained convolutional neural network enhances the images acquired by a smartphone-based microscope. The network's output images correct spectral and spatial aberrations, blocking artifacts, and achieve increased SNR and spatial resolution. Top row: lung tissue section. Bottom row: blood smear.

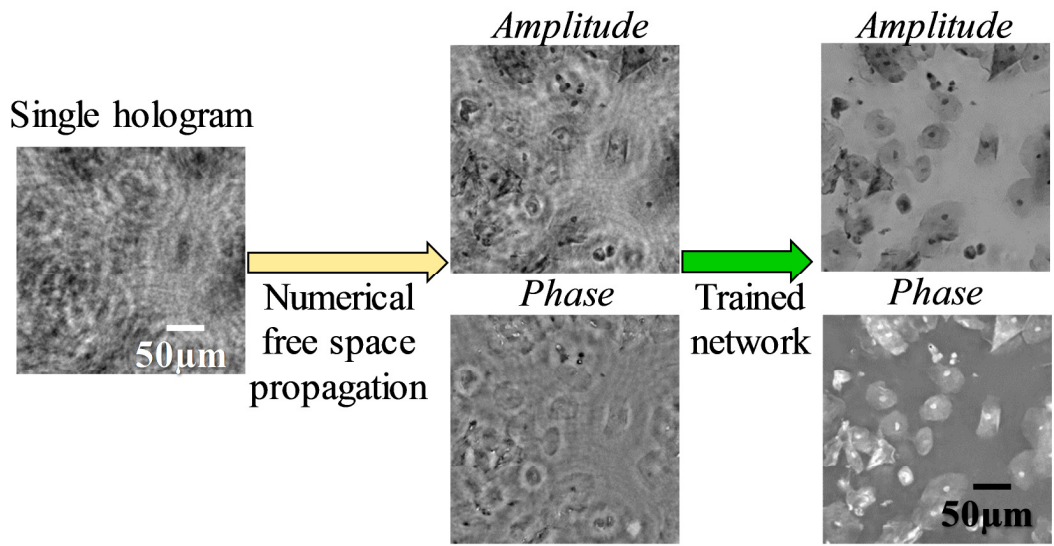

**Fig. 4.** Holographic image reconstruction and phase recovery using a deep neural network. Due to the missing phase information, the object's complex-valued image is distorted by the twin image and self-interference artifacts, following the free space back propagation without phase (middle column). The deep network is trained to reconstruct the true object distribution (both phase and amplitude), non-iteratively and using a single hologram of the sample. The sample shown here is a Pap smear.